\documentclass[pdflatex,sn-mathphys-num]{sn-jnl}

\usepackage{graphicx}%
\usepackage{multirow}%
\usepackage{amsmath,amssymb,amsfonts}%
\usepackage{amsthm}%
\usepackage{mathrsfs}%
\usepackage[title]{appendix}%
\usepackage{xcolor}%
\usepackage{textcomp}%
\usepackage{hyperref}
\usepackage{manyfoot}%
\usepackage{booktabs}%
\usepackage{algorithm}%
\usepackage{algorithmicx}%
\usepackage{algpseudocode}%
\usepackage{listings}%
\usepackage{svg}
\usepackage{tabularx}
\usepackage{float}
\usepackage{enumitem}
\usepackage{colortbl}
\usepackage{xcolor}
\usepackage{pgf}

\definecolor{lightred}{rgb}{1,0.8,0.8}
\definecolor{medred}{rgb}{1,0.5,0.5}
\definecolor{darkred}{rgb}{1,0.2,0.2}

\setlist[itemize]{left=1em, label=--, topsep=1pt, itemsep=1pt}

\theoremstyle{thmstyleone}%
\theoremstyle{thmstyletwo}%

\theoremstyle{thmstylethree}%

\begin{document}

\title[Before the Outrage: Challenges and Advances in Predicting Online Antisocial Behavior]{Before the Outrage: Challenges and Advances in Predicting Online Antisocial Behavior}


\author*[1]{ \sur{Anaïs Ollagnier}}\email{ollagnier@i3s.unice.fr}



\affil*[1]{ \orgname{Université Côte d’Azur, Inria, CNRS, I3S}, \orgaddress{\street{930 route des Colles, BP 145}, \city{Sophia Antipolis Cedex}, \postcode{06903}, \country{France}}}




\abstract{Antisocial behavior (ASB) on social media—including hate speech, harassment, and trolling—poses growing challenges for platform safety and societal well-being. While prior work has primarily focused on detecting harmful content after it appears, predictive approaches aim to forecast future harmful behaviors—such as hate speech propagation, conversation derailment, or user recidivism—before they fully unfold. Despite increasing interest, the field remains fragmented, lacking a unified taxonomy or clear synthesis of existing methods. This paper presents a systematic review of over 49 studies on ASB prediction, offering a structured taxonomy of five core task types: early harm detection, harm emergence prediction, harm propagation prediction, behavioral risk prediction, and proactive moderation support. We analyze how these tasks differ by temporal framing, prediction granularity, and operational goals. In addition, we examine trends in modeling techniques—from classical machine learning to pre-trained language models—and assess the influence of dataset characteristics on task feasibility and generalization. Our review highlights methodological challenges, such as dataset scarcity, temporal drift, and limited benchmarks, while outlining emerging research directions including multilingual modeling, cross-platform generalization, and human-in-the-loop systems. By organizing the field around a coherent framework, this survey aims to guide future work toward more robust and socially responsible ASB prediction.}




\maketitle

\section{Introduction}

Social media platforms have profoundly transformed how individuals consume information, engage in public discourse, and build communities online~\cite{DBLP:journals/nms/KahnK04,brown2007word,Quattrociocchi2014Opinion}. Their microblogging infrastructure and decentralized architecture allow users to participate in diverse conversations at an unprecedented scale~\cite{DBLP:journals/giq/CriadoSG13,DBLP:journals/corr/abs-2211-15988}, amplifying both civic engagement and the rapid spread of ideas~\cite{DBLP:journals/pnas/CinelliMGQS21}. However, these same affordances have also created fertile ground for harmful dynamics. Research has consistently shown that social media platforms facilitate the spread of misinformation~\cite{DBLP:journals/corr/VicarioVBZSCQ16}, foster ideological echo chambers~\cite{DBLP:journals/pnas/CinelliMGQS21}, and escalate hostility in online interactions~\cite{DBLP:conf/cscw/ChengBDL17,DBLP:conf/socinfo/QuattrociocchiE22,DBLP:conf/www/SaveskiRR21}.

These hostile environments often give rise to diverse forms of antisocial behavior (ASB)—including cyberbullying~\cite{DBLP:conf/flairs/OllagnierCV23}, hate speech~\cite{DBLP:journals/snam/OllagnierCV23}, trolling~\cite{doi:10.1177/0267323118760323}, and sexual harassment~\cite{DBLP:conf/acl/ChowdhurySSM19}. Such behaviors carry severe consequences for individuals and societies alike, including psychological distress, polarization, and loss of trust in digital spaces~\cite{Parent2019Social,DBLP:conf/websci/SahaCC19,DBLP:journals/isci/ValensiseCQ23}. As platforms struggle to scale moderation and accountability, there is a growing need for predictive systems that can anticipate harmful interactions—enabling earlier, more effective interventions before toxicity spreads or escalates.

While detection of harmful content has been extensively studied~\cite{DBLP:journals/lre/PolettoBSBP21,DBLP:journals/scientometrics/TontodimammaNSF21,DBLP:journals/ijon/Jahan023}, ASB prediction—defined as the computational modeling of future harmful behaviors or outcomes—remains comparatively underexplored and methodologically fragmented. Predictive tasks span diverse goals, such as forecasting harassment intensity~\cite{dahiya2021would,meng2023predicting}, predicting whether a conversation will derail~\cite{DBLP:journals/corr/abs-2306-12982,yuan2023conversation}, or identifying potential hate spreaders before they act~\cite{irani2021early}. These tasks differ in temporal scope (e.g., ex-ante vs. peeking strategies), analytical focus (e.g., micro-, meso-, or macro-level), and formulation (e.g., classification or regression). Moreover, they are often tightly coupled to the platform on which the data are collected, limiting the generalizability of findings and the comparability of methods across studies.

To address this gap, this survey provides a systematic review of ASB prediction research, offering a unified framework for analyzing prediction goals, modeling strategies, and evaluation protocols. We specifically focus on forward-looking tasks that go beyond static detection to support proactive moderation and risk-aware design. By mapping how ASB prediction has evolved across platforms, feature types, and learning paradigms, we aim to clarify the current landscape and highlight opportunities for broader, more robust solutions.

Our review is guided by the following research questions:
\begin{itemize}
    \item \textbf{RQ1:} What types of predictive tasks are addressed in current ASB literature? This includes the outcomes modeled (e.g., escalation, spread, recurrence), temporal framing, and practical objectives.
    
    \item \textbf{RQ2:} What modeling techniques are used in ASB prediction, and how have they evolved? We consider traditional machine learning, neural models, pre-trained language models, and multimodal or structured architectures, along with the types of features they rely on (e.g., content-based, behavioral, contextual, temporal).
    
    \item \textbf{RQ3:} What types of data are used for ASB prediction, and how do dataset characteristics influence task feasibility and generalization? This includes platform specificity, data structure (e.g., conversational flow vs. network spread), and the emergence of task-specific and multilingual corpora.
\end{itemize}

\textbf{Contributions.} This survey makes the following contributions:
\begin{itemize}
    \item We propose a unified taxonomy of ASB prediction tasks that captures temporal, structural, and behavioral dimensions.
    \item We analyze over 49 representative studies, linking their predictive goals, features, models, and evaluation strategies.
    \item We assess methodological trends and recurring challenges in ASB prediction, such as dataset scarcity, temporal drift, limited benchmarks, and the trade-offs between model performance and interpretability.
    \item We highlight emerging needs—including multilingual and cross-platform modeling, human-in-the-loop design, and shared evaluation tasks—that define future directions in ASB prediction research.
\end{itemize}

\textbf{Paper structure.} The remainder of the paper is organized as follows: Section~\ref{sec:background} defines ASB prediction and situates it within related areas as well as  introduces our task taxonomy. Section~\ref{sec:sota} describes the review methodology. Section~\ref{sec:slr} and~\ref{sec:analysis} synthesize findings on prediction tasks, modeling approaches, and dataset use. Section~\ref{sec:challenges} discusses open challenges and opportunities. We conclude in Section~\ref{sec:conclusion} with a summary and future directions.

\section{Background}
\label{sec:background}
To contextualize ASB prediction, we survey definitions of antisocial behavior in the literature, outline prevalent modeling approaches, and situate them within related computational tasks.

\subsection{Definitions}
\label{sec:definition}

This study centers on two foundational concepts: \textit{antisocial behavior} and \textit{prediction}. In computational research, harmful online interactions have been described using a variety of overlapping terms, including \textit{hate speech}, \textit{online abuse}, \textit{toxic language}, and \textit{cyberaggression}~\cite{DBLP:journals/lre/PolettoBSBP21,DBLP:journals/information/AlkomahM22,DBLP:journals/ijon/Jahan023}. While these terms are often used interchangeably, they differ significantly in scope, intent, and interpretability—both legally and socially—which presents challenges for consistent problem formulation and predictive modeling.

Historically, research in this area has centered on the concept of \textit{hate speech}, a term widely adopted across academic, legal, and policy discourse. However, its usage is often inconsistent, frequently serving as a catch-all label for a broad range of offensive or harmful content. This lack of terminological precision has led to conceptual ambiguity and challenges in task formulation. In response, more nuanced definitions have been proposed. For example, Poletto et al.~\cite{DBLP:journals/lre/PolettoBSBP21} define hate speech as a specific subcategory of abusive language, distinguished by two key criteria: (1) its function—such as inciting violence or threatening dignity or safety—and (2) its target—namely individuals or groups attacked based on identity characteristics rather than personal behavior. This narrower framing enhances alignment with human rights norms and legal frameworks, but also underscores the broader inconsistency in how harmful online behavior is defined and operationalized across studies.

Still, hate speech captures only a fraction of the behaviors considered harmful in online settings. To address the broader landscape, we adopt the term \textit{antisocial behavior (ASB)} as an overarching category that encompasses a wide spectrum of harmful actions in digital environments. Drawing from social psychology, platform policy guidelines (e.g., Facebook Community Standards~\cite{meta2022community}), and recent work in computational social science~\cite{gruzd2020studying, doi:10.1177/20563051231196874}, we distinguish three primary types of ASB:

\begin{itemize}
\item \textbf{Personal harms} — Actions directed at individuals, such as cyberbullying or targeted harassment.
\item \textbf{Group-directed harms} — Attacks based on group identity, including hate speech and stereotyping.
\item \textbf{Environmental disruptions} — Behaviors that degrade the health of online communities, such as trolling, spamming, misinformation, coordinated raids, or evasion of moderation.
\end{itemize}

This classification emphasizes not only the form and severity of harmful behaviors but also their social targets and contextual dynamics. ASB includes both \textit{linguistic content} (e.g., incivility, toxicity) and \textit{behavioral patterns} (e.g., repeated provocations, manipulation tactics), and often unfolds over time or through interaction. Unlike narrower definitions that focus exclusively on single messages or comments, the ASB framing accommodates temporal aspects, escalation patterns, relational context, and platform-specific dynamics. By conceptualizing ASB in this way, we enable a more holistic approach to prediction—one that seeks not just to detect harmful content, but to anticipate \textit{when}, \textit{how}, and \textit{through whom} such behaviors are likely to arise or recur.
\vspace{0.5em}

ASB prediction refers to a class of computational tasks aimed at anticipating the likelihood, trajectory, or impact of harmful behaviors in online environments. These tasks vary widely in terms of objectives, granularity, and timing. Key task formulations include:

\begin{itemize}
\item \textbf{Binary or multiclass classification}: e.g., Will a comment thread turn toxic?
\item \textbf{Regression}: e.g., How likely is a user to reoffend after a ban?
\end{itemize}

The temporal dimension plays a pivotal role in ASB prediction, shaping both task design and modeling strategy. Some approaches focus on \textit{early prediction}, aiming to identify potential harm based on partial signals---such as the first few messages in a conversation---before escalation occurs. Others explore \textit{proactive prediction}, where harm is anticipated even before content is posted, enabling pre-publication moderation or real-time intervention. This direction aligns closely with emerging goals in platform safety and governance. By contrast, \textit{retrospective modeling}---which analyzes complete interaction histories to understand harmful behavior after it has occurred---falls outside the scope of this study. While valuable for behavioral profiling and forensic analysis, such approaches do not align with our focus on the \textit{computational modeling of future harmful behaviors or outcomes}. Another emerging yet underdeveloped task is \textit{diffusion prediction}, which estimates how harmful content may spread through user networks. Though widely applied in domains like viral marketing and information cascades~\cite{DBLP:conf/www/ChengADKL14,DBLP:conf/cikm/CaoSCOC17,yatish2020recent}, its use in ASB---such as forecasting the virality of hate speech or identifying likely amplifiers of toxic content---remains a developing frontier~\cite{DBLP:journals/csur/ZhouXTZ21}. Together, these challenges highlight the complexity of ASB prediction. It is not a singular problem, but a \textit{diverse set of temporally and operationally distinct tasks} that vary in scope, data availability, modeling goals, and societal implications---necessitating a unified taxonomy to map the field and guide future work.

\subsection{Taxonomy}
\label{sec:taxonomy}

There is currently no unified taxonomy that comprehensively describes the range of computational tasks addressed in the prediction of ASB online. To address this gap, we derive a taxonomy grounded in a systematic review of recent literature (Section~\ref{sec:slr}), focusing on how predictive tasks are framed in relation to the lifecycle of antisocial behaviors. We define \textit{antisocial behavior prediction} as the computational modeling of future harmful online behaviors—whether they are incipient, ongoing, or likely to recur. These behaviors include, but are not limited to, hate speech, harassment, trolling, cyberbullying, and coordinated toxic campaigns~\cite{DBLP:conf/icwsm/ChengDL15,gruzd2020studying, doi:10.1177/20563051231196874}. Our proposed taxonomy organizes the space of predictive tasks along two key dimensions:
\begin{enumerate}
    \item \textbf{Temporal orientation} — whether the model anticipates the emergence, escalation, or spread of harm.
    \item \textbf{Operational purpose} — whether the model is intended to support moderation, risk assessment, or intervention.
\end{enumerate}

The result is five distinct yet complementary categories that reflect common predictive goals across studies. Each category is associated with specific modeling challenges, types of input data, and intervention strategies. Table~\ref{tab:taxonomy} provides an overview of the taxonomy along with representative examples.

\begin{table}[h]
\centering
\footnotesize
\caption{Taxonomy of antisocial behavior prediction tasks, with descriptions and representative examples.}
\label{tab:taxonomy}
\begin{tabular}{|p{2.5cm}|p{9.5cm}|}
\hline
\textbf{Category} & \textbf{Definition, Focus, and Example Tasks} \\ \hline

\textbf{Harm Emergence} &
Forecast whether an ongoing but currently civil or neutral interaction will develop into harmful behavior. \textit{Output:} A prediction (often binary or multi-class) about future harm development. \\
& \begin{tabular}[t]{@{}l@{}}
-- Will a tweet’s replies eventually contain abuse?\\
-- Will a Wikipedia discussion derail into personal attacks?\\
-- Will hate speech emerge in a Reddit thread that is currently civil?
\end{tabular} \\ \hline

\textbf{Harm Propagation} &
Predict how far, how fast, or through whom harmful content will spread once it already exists. \textit{Output:} A prediction of diffusion scope or intensity (binary, count-based, or continuous). \\
& \begin{tabular}[t]{@{}l@{}}
-- Will a toxic post go viral (liked/retweeted)?\\
-- Which users will reshare harmful narratives?\\
-- How deep will a trolling cascade reach in the network?
\end{tabular} \\ \hline

\textbf{Behavioral Risk} &
Assess the likelihood that a particular user will engage in, or become a target of, harmful behavior. \textit{Output:} A user-level prediction (e.g., risk scores, binary labels). \\
& \begin{tabular}[t]{@{}l@{}}
-- Will a user reoffend after a ban?\\
-- Is a user likely to adopt or spread hate ideologies?\\
-- Will a user migrate to more extreme communities?
\end{tabular} \\ \hline

\textbf{Early Harm Detection} &
Identify harmful behavior at its very onset, based on minimal early signals (e.g., first few comments). \textit{Output:} A rapid prediction of imminent harm to trigger early intervention. \\
& \begin{tabular}[t]{@{}l@{}}
-- Flag bullying after the first two posts in a thread.\\
-- Detect signs of personal attacks from initial comments.
\end{tabular} \\ \hline

\textbf{Proactive Moderation} &
Support preventive action by evaluating content \textit{before} it is posted or fully shared. \textit{Output:} A pre-publication prediction or suggestion to reduce risk. \\
& \begin{tabular}[t]{@{}l@{}}
-- Flag risky content at posting time.\\
-- Recommend safer phrasing for a comment draft.\\
-- Rank new posts by potential harm for moderator review.
\end{tabular} \\ \hline

\end{tabular}
\end{table}

\paragraph{Harm Emergence Prediction.}

This task involves forecasting whether an initially civil or benign interaction—such as a forum thread, comment exchange, or reply chain—will eventually develop into harmful behavior. The core objective is to detect subtle warning signs that precede toxicity, such as rhetorical escalation, negative sentiment shifts, or social power dynamics. Unlike harm detection tasks that identify toxicity already present, harm emergence prediction anticipates its future appearance before any explicit harmful content is visible. For instance, models may predict whether a Wikipedia Talk page or Reddit CMV discussion will transition from constructive disagreement to personal attacks or insults. These models typically operate at the thread level, analyzing the sequential development of conversations to determine if a derailment is imminent.

\paragraph{Harm Propagation Prediction.}

This category addresses how harmful content spreads across a network once it exists, focusing on its diffusion dynamics rather than its initial emergence. Models in this space estimate the scale, speed, or directionality of toxic content amplification. Tasks often include predicting whether a hateful or inflammatory post will go viral, identifying which users are most likely to reshare or endorse the content, or determining how far and through which communities it may propagate. Techniques commonly draw from network science and diffusion modeling, considering both structural features (e.g., follower networks, cascade depth) and temporal indicators (e.g., rate of resharing). 

\paragraph{Behavioral Risk Prediction.}

Behavioral risk prediction focuses on individuals rather than specific content, aiming to assess whether a user is likely to engage in or become a target of antisocial behavior. These models build risk profiles using past user behavior, engagement patterns, linguistic tendencies, and contextual signals. They are commonly used to predict recidivism after a moderation action, detect users at risk of adopting toxic ideologies, or identify individuals vulnerable to targeted harassment. For example, one might predict whether a user who has posted in a hate subreddit is likely to migrate to other toxic communities. The predictions can be used to inform preemptive moderation, user interventions, or safety-focused design adjustments.

\paragraph{Early Harm Detection.}

Early harm detection emphasizes speed and minimal input, seeking to identify harmful interactions in their infancy—often after just the first few posts or turns in a conversation. Unlike harm emergence prediction, which focuses on the developmental arc toward toxicity, early detection prioritizes the earliest feasible intervention point. This allows systems to flag conversations at risk of becoming toxic based on only limited context, such as the first two or three comments in a thread. These models are particularly useful in high-volume settings, where rapid triage is critical. Early detection enables timely moderation and helps prevent escalation, even before patterns of toxicity fully unfold.

\paragraph{Proactive Moderation Support.}

Proactive moderation support encompasses a suite of models and tools designed to assist moderators or users in preventing harm before it occurs. This includes predicting the potential harmfulness of content at the time of authoring or submission, recommending edits to reduce inflammatory tone, or ranking posts by their likelihood of needing review. Such models may be embedded in content creation interfaces or used in human-in-the-loop moderation workflows. The goal is not just to detect or respond to harm, but to support safer communication by nudging users or systems toward preemptive action. For example, a platform might suggest less aggressive phrasing to a user composing a contentious comment or prioritize certain flagged posts for moderator attention.

\vspace{0.5em}
Together, these five categories reflect a shift in the field from reactive moderation toward anticipatory and preventive approaches. By structuring ASB prediction tasks along these axes, the proposed taxonomy provides a conceptual foundation for understanding and comparing methods, identifying research gaps, and designing systems that better align with the needs of digital safety and governance.

\section{Methodology for Collecting Related Review Papers}
\label{sec:sota}

\begin{figure}[ht]
\centering
  \includegraphics[width=\linewidth]{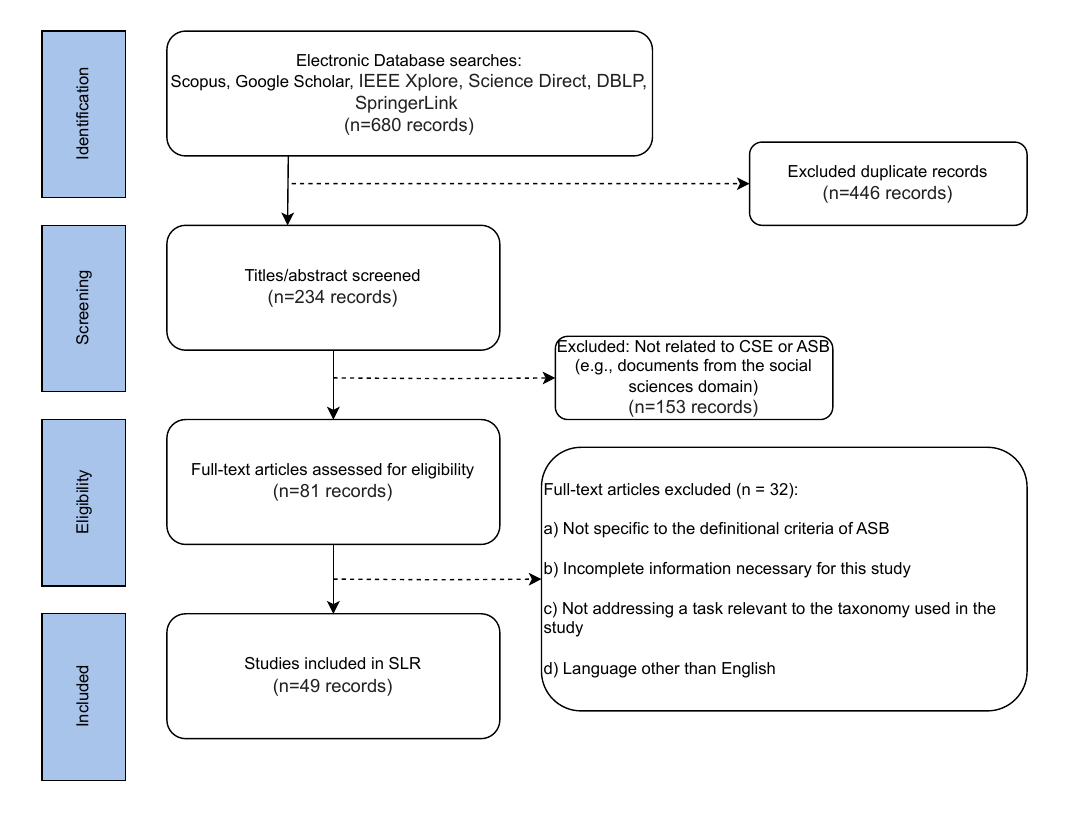}
\caption{PRISMA flowchart for selection of antisocial behaviour prediction studies.}
\label{fig:prisma}
\end{figure}

While tasks related to the detection or identification of hate speech and cyberbullying have been extensively studied, the prediction of these phenomena—along with broader forms of ASB—remains a relatively new and emerging area. In compiling this survey, we adopted a systematic review-based approach, drawing on the guidelines proposed by~\citet{kitchenham2004systematic} for conducting systematic reviews in computer science and engineering (CSE), while adapting them to the specific context of ASB research. Additionally, our methodology follows the PRISMA framework~\cite{moher2009preferred}, which is widely used in systematic reviews to ensure transparency and reproducibility. The study selection process is summarized in Fig.~\ref{fig:prisma}. This section outlines the main steps undertaken in the research process.

\subsection{Search Keywords}

To collect relevant papers on ASB prediction, we first identified and selected representative keywords to retrieve pertinent publications from major scholarly databases. Given that ASB prediction is a relatively recent and developing research topic, we broadened our search scope to include terms reflecting both general forms of ASB and specific aspects of predictive modeling. The final set of search terms included terminology related to predictive tasks—such as \textit{prediction}, \textit{forecasting}, \textit{early detection}, \textit{propagation}, \textit{diffusion}, \textit{temporal modeling}, and \textit{real-time prediction}—combined with concepts associated with online harm, including \textit{antisocial behaviour}, \textit{cyberaggression}, \textit{hate speech}, \textit{cyberbullying}, \textit{online abuse}, \textit{online harassment}, \textit{toxicity}, \textit{online grooming}, and \textit{hate crime}. These keywords were used in various logical combinations to ensure broad coverage and maximize retrieval of relevant literature.

\subsection{Sources}

We collected peer-reviewed academic works from a wide range of scholarly sources, including Google Scholar, DBLP, Scopus, IEEE Xplore, SpringerLink, and ScienceDirect. For the latter group of databases, we utilized the content aggregator tool \texttt{SciLEx}\footnote{\url{https://github.com/datalogism/SciLEx}} to streamline and centralize the search process. The defined keywords were applied to titles and abstracts to identify publications relevant to ASB prediction. To ensure relevance and quality, we applied two filtering criteria: (1) a publication date range from 2010 to 2025, and (2) a language restriction, including only works written in English. The final search was conducted on June 17, 2025. For each retrieved record, we extracted key metadata, including the title, abstract, authors’ names and affiliations, journal name, and year of publication. These records were compiled into a structured spreadsheet for further screening and analysis.

\subsection{Inclusion and Exclusion Criteria}

All works not directly related to ASB—including those focused on narrowly scoped or tangential subjects, those not employing an ML-based approach, and those that were either not peer-reviewed or not openly accessible—were excluded from our review. One of the major challenges we encountered was the blurred boundary between prediction and other computational tasks such as detection, classification, identification, recognition, analysis, and profiling. The absence of a unified conceptual framework in the domain leads to inconsistencies in terminology and task definitions: the same task may be described using different labels across studies, while broad terms like ``detection'' or ``prediction'' are frequently applied to tasks that share core computational goals. Accordingly, we include only studies that employ binary or multi-class classification or regression techniques and explicitly aim to forecast one or more of the following: the early signals of harm, the emergence of antisocial behavior, its potential propagation, the behavioral risk posed by users, or outcomes relevant to proactive moderation and intervention strategies. Studies that did not meet these inclusion criteria were systematically excluded.

\subsection{Data Extraction and Thematic Synthesis}

The initial screening phase involved reading titles and abstracts to filter out clearly irrelevant results. Articles that passed this preliminary assessment were subsequently reviewed in full to confirm their alignment with the inclusion criteria. Duplicate records—retrieved from multiple databases—were identified and removed to avoid redundancy. Following the screening process, we carried out structured data extraction and synthesis. Each eligible paper underwent a full-text review and was annotated using a custom-designed classification scheme. This scheme captures several key dimensions of each study, including its disciplinary grounding (e.g., social computing, natural language processing, or computational social science), primary research contribution (e.g., predictive modeling, resource development, or intervention design), methodological approach (e.g., neural architectures, traditional machine learning, or hybrid systems), and the specific type of ASB prediction task addressed—such as forecasting harm emergence, behavioral risk, or content propagation—based on the taxonomy introduced in Section~\ref{sec:taxonomy}. Throughout the review process, citation management software (Zotero) was used to manage references and maintain consistency across the data. To support the synthesis of findings, we applied thematic analysis~\cite{kitchenham2004systematic}, a qualitative technique well-suited for identifying patterns in textual data. Through iterative close reading and content coding, we extracted information relevant to our research questions and organized it into thematically coherent categories. These codes were then abstracted into higher-order themes, enabling us to identify and articulate a set of key challenges currently facing the field of ASB prediction.

In total, this process yielded \textbf{49} papers that met all inclusion criteria and were included in the final analysis. These papers provide the foundation for our literature synthesis and inform our discussion of the current state of research and open challenges in ASB prediction.

\section{Analysis of Retrieved Literature}
\label{sec:slr}
This section presents the key findings from our analysis of the retrieved literature on ASB prediction. We first explore the evolution of publication activity, then highlight seminal contributions, followed by an analysis of the field’s conceptual structure.

\subsection{Research activity}

\begin{figure}[ht]
\centering
\includegraphics[width=\linewidth]{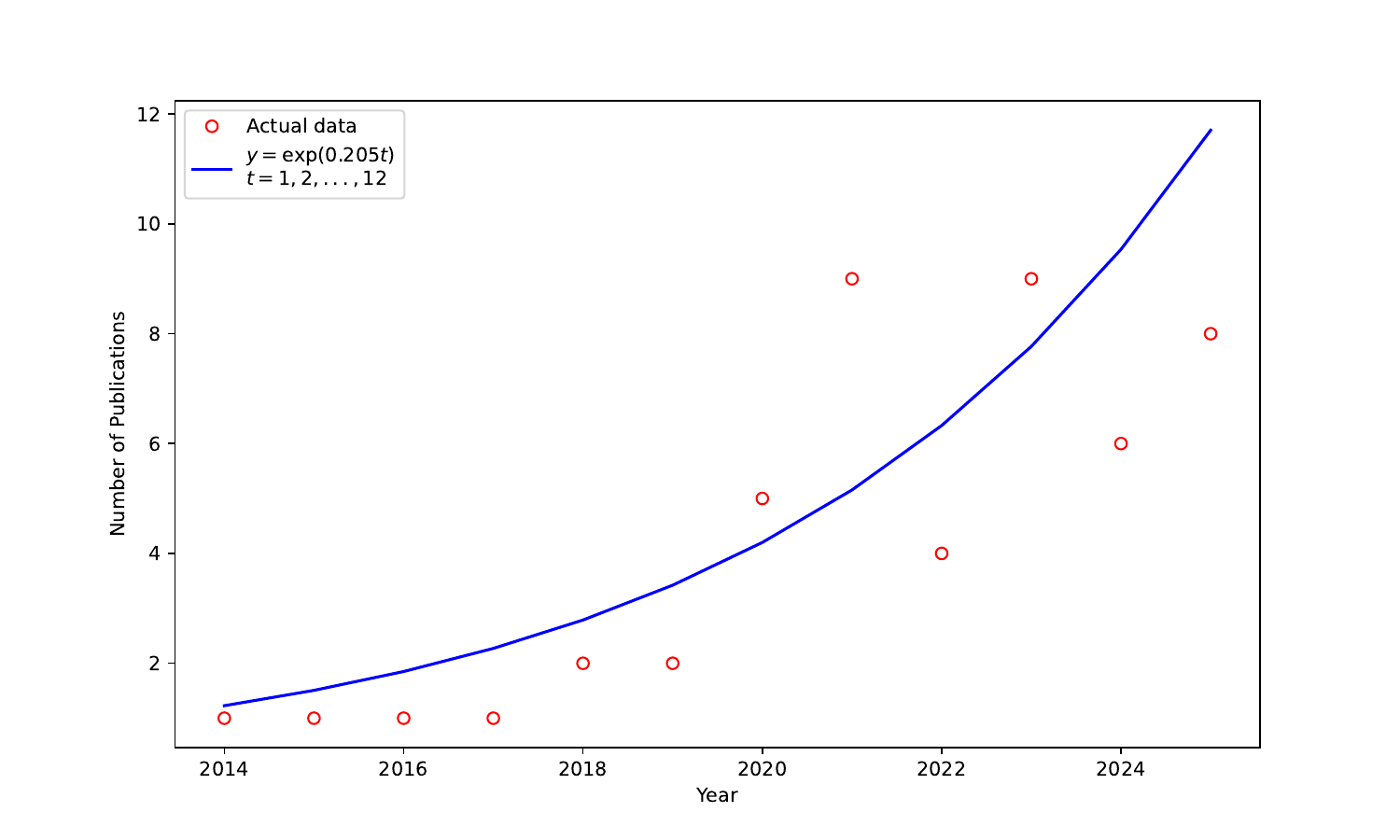}
\caption{Number of publications on antisocial behaviour prediction per year: observed and expected distribution.}
\label{fig:years}
\end{figure}


The temporal distribution of publications in the field of ASB prediction reveals a clear and accelerating growth trajectory over the past decade. As illustrated in Fig.~\ref{fig:years}, the early years (2014--2017) were characterized by sporadic contributions---just one publication per year---highlighting the field’s nascent and exploratory phase. A modest increase in activity emerged in 2018 and 2019, with two publications each year, signaling the beginning of broader scholarly interest. A more substantial rise occurred in 2020, with five publications, followed by a sharp increase in 2021. That year, along with 2023, marked the peak of productivity, each with nine publications. This surge reflects growing momentum driven by heightened awareness of online harms, improved access to social media data (via APIs or platform datasets), and rapid advances in natural language processing and machine learning. It also coincides with growing regulatory pressure---most notably, the EU’s integration of its Hate Speech Code of Conduct into the legally binding Digital Services Act\footnote{\href{https://digital-strategy.ec.europa.eu/en/library/code-conduct-countering-illegal-hate-speech-online}{EU Digital Services Act}} in early 2025, as well as global initiatives such as the UN’s International Day for Countering Hate Speech\footnote{\href{https://www.un.org/en/hate-speech/un-strategy-and-plan-of-action-on-hate-speech}{UN Strategy and Plan of Action on Hate Speech}} and the \#IamHere counter-speech movement\footnote{\href{https://iamhereinternational.com/}{\#IamHere International}}. These developments likely reinforced both policy relevance and research interest in predictive moderation systems. The upward trend continues into 2024, which saw six publications, and 2025, which already counts eight publications despite the year being incomplete at the time of review. Although 2022 saw a slight dip, with four publications, it remains consistent with the overall pattern of expansion. Importantly, this growth is not limited to publication quantity alone; it is accompanied by increasing diversity and prestige in publication venues. Since 2020, there has been a notable proliferation of research across top-tier conferences and journals. The Web Conference, for example, appears regularly, indicating sustained engagement with the web and social media research community. High-impact venues such as EMNLP, NAACL, and ICWSM reflect the field’s strong alignment with cutting-edge developments in natural language processing and computational social science. Meanwhile, venues like KDD, \emph{IEEE Transactions on Network Science and Engineering}, and \emph{Future Generation Computer Systems} point to deep technical engagement and interdisciplinary relevance.
Together, these trends indicate not only a rise in research activity but also a deepening of the field’s methodological and conceptual sophistication. This evolution signals a growing recognition of the importance of anticipating and mitigating ASB before it escalates—an imperative now addressed through increasingly data-driven and interdisciplinary approaches.


\subsection{Major publications}

\begin{table}[ht]
\centering
\caption{Survey of Key Publications on Antisocial Behavior Prediction}
\label{tab:survey_reviews}
\begin{tabular}{@{}p{4.4cm}p{2.5cm}p{1.5cm}p{2.2cm}p{0.8cm}@{}}
\toprule
\textbf{Paper Title} & \textbf{Authors, Year} & \textbf{Venue} & \textbf{Prediction Task} & \textbf{Cites} \\
\midrule
\textit{Antisocial Behavior in Online Discussion Communities} & \citet{DBLP:conf/icwsm/ChengDL15}, 2015 & ICWSM & Behavioral Risk Prediction & 323 \\[3pt]
\textit{Prediction of Cyberbullying Incidents in a Media-Based Social Network} & \citet{DBLP:conf/asunam/HosseinmardiRHL16}, 2016 & ASONAM & Harm Emergence Prediction & 88 \\[3pt]
\textit{Constructing Interval Variables via Faceted Rasch Measurement and Multitask Deep Learning} & \citet{DBLP:journals/corr/abs-2009-10277}, 2020 & arXiv & Behavioral Risk Prediction & 87 \\[3pt]
\textit{The Structure of Toxic Conversations on Twitter} & \citet{DBLP:conf/www/SaveskiRR21}, 2021 & WWW & Early Harm Detection & 85 \\[3pt]
\textit{Cyberbullying Detection Using Time Series Modeling} & \citet{DBLP:conf/icdm/PothaM14}, 2014 & ICDM WS & Early Harm Detection & 63 \\
\bottomrule
\end{tabular}
\end{table}

Table~\ref{tab:survey_reviews} presents a curated selection of influential publications that have shaped the predictive modeling of ASB in online environments. These works span key subdomains such as behavioral risk prediction, harm emergence forecasting, and early detection, showcasing both methodological diversity and the evolution of research goals—from reactive content moderation to proactive, user- and context-aware forecasting.

A foundational contribution by~\citet{DBLP:conf/icwsm/ChengDL15} introduced one of the first large-scale analyses of antisocial users across multiple communities. By examining behavioral dynamics over time, they demonstrated that harmful user behavior is often detectable before bans are enforced. Their work revealed that concentrated activity, low relevance, and escalating community rejection are predictive markers. Moreover, they showed that negative community feedback loops can exacerbate user misconduct, highlighting the need for early intervention models that account for social context. Building on this preventive paradigm,~\citet{DBLP:conf/asunam/HosseinmardiRHL16} proposed an anticipatory model for cyberbullying prediction on Instagram, leveraging multimodal features from the initial post stage—including caption text (e.g., profanity indicators), image embeddings, and ego-network features. Their approach marked a shift from detection to pre-incident forecasting, achieving strong predictive performance on real-world data and demonstrating the feasibility of using pre-comment signals for proactive cyberbullying mitigation. In a more technically sophisticated approach,~\citet{DBLP:journals/corr/abs-2009-10277} introduced a continuous measurement framework for hate speech severity. They combined faceted Rasch item response theory with multitask deep learning, modeling hate as a debiasable, interpretable latent variable composed of multiple constituent components. The final neural predictor mapped text to continuous scores using ordinal loss functions and Transformer-based encoders (e.g., BERT, RoBERTa). This model notably outperformed the Perspective API, offering a scalable and explainable alternative for fine-grained behavioral risk prediction across platforms. A complementary line of research by~\citet{DBLP:conf/www/SaveskiRR21} investigated the predictive role of conversational structure in online toxicity. Using a dataset of over 58 million tweets from politically sensitive conversations, they showed that features like reply tree depth, breadth, and social graph sparsity are predictive of future toxicity. Their framework enabled early prediction at both the thread and user-reply levels, and their experiments demonstrated that combining structural and linguistic features enhances performance—supporting the use of conversation dynamics for scalable, real-time moderation. Finally,~\citet{DBLP:conf/icdm/PothaM14} introduced a time series-based approach for forecasting the severity of cyberbullying in predator–victim conversations. They transformed dialogue sequences into SVD-reduced signals and applied Dynamic Time Warping to detect correlations between the linguistic progression and attack severity. By training neural classifiers over these temporal embeddings, the study enabled forecasting based on partial conversation history, advancing the field beyond static text classification toward sequential behavior modeling.

Together, these contributions illustrate a broadening of the ASB prediction landscape—from user profiling and content scoring to diffusion modeling and conversational dynamics—marking a shift toward early, explainable, and context-sensitive interventions in online harm prevention.


\subsection{Conceptual structure of antisocial behaviour prediction research}

\begin{figure}[ht]
  \centering
  \includegraphics[scale=0.4]{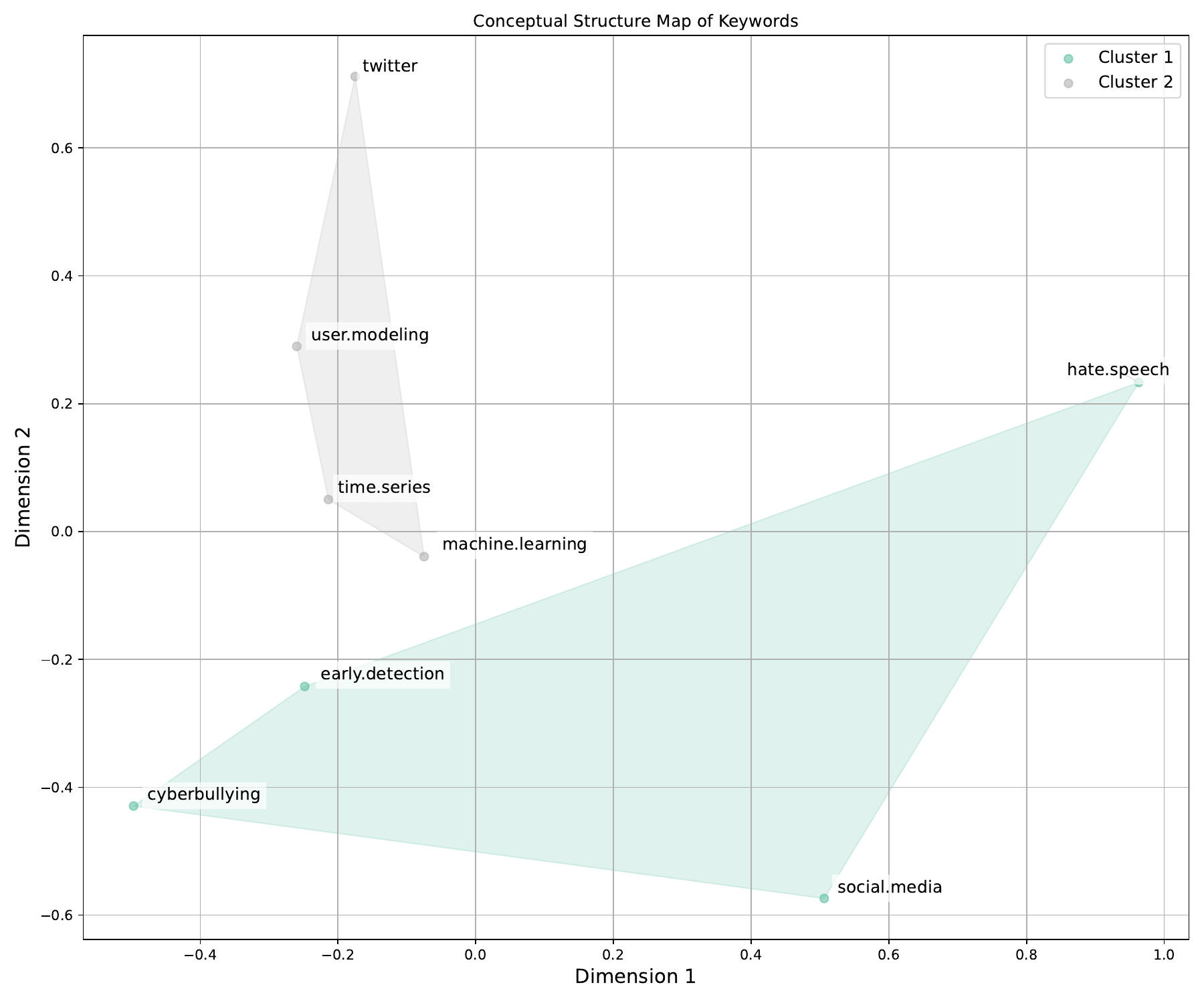}
  \caption{Conceptual map of antisocial behaviour prediction research.}
  \label{fig:mca}
\end{figure}

The conceptual structure of research on ASB prediction is illustrated in Fig.~\ref{fig:mca}, where authors’ keywords—occurring more than five times—are projected onto a two-dimensional plane generated through Multiple Correspondence Analysis (MCA).

The conceptual structure map provides a visual synthesis of the research landscape surrounding ASB prediction, organizing the domain into two distinct thematic clusters. Cluster 1 centers on the core social and behavioral aspects of ASB. It includes keywords such as cyberbullying, early detection, hate speech, and social media. These terms emphasize the phenomenological and application-driven focus of the research, where the goal is to understand and intervene in harmful online behaviors as they emerge in social contexts. The inclusion of early detection highlights a proactive approach, while the presence of social media and hate speech points to platform-based studies where discourse analysis and societal impact are key concerns. Cluster 2, by contrast, is more technically oriented, composed of terms such as machine learning, time series, Twitter, and user modeling. This grouping reflects research focused on methodological innovation and predictive modeling, especially using data-driven approaches to forecast user behavior or content dynamics over time. The prominence of Twitter and user modeling suggests a focus on micro-level interactions and user-specific features, while time series analysis supports the development of temporally-aware predictive frameworks. The map reveals a complementary relationship between the two clusters. Cluster 1 grounds the field in social relevance and real-world issues, while Cluster 2 provides the computational tools and models that enable effective prediction.




\section{Landscape of Antisocial Behavior Prediction: Tasks, Techniques, and Data}
\label{sec:analysis}

In this section, we address the three research questions that structure our review of the literature on ASB prediction:
\begin{itemize}
    \item \textbf{RQ1 – Predictive Tasks:} How are predictive approaches to ASB differentiated in the literature by task type, timing, and granularity?
    \item \textbf{RQ2 – Methods:} What machine learning approaches are used for ASB prediction, and how have they evolved over time?
    \item \textbf{RQ3 – Datasets:} How do dataset sources and structures shape ASB prediction tasks in the literature?
\end{itemize}

For each research question, we examine key strategies and trends, drawing on both the taxonomy introduced in Section~\ref{sec:taxonomy} and established dimensions of information modeling to highlight recurring patterns~\cite{DBLP:journals/csur/ZhouXTZ21}, methodological approaches, and gaps across the field of ASB prediction.

\subsection{RQ1 - Characterizing Predictive Task Types in ASB Detection}

To better characterize the multifaceted nature of ASB prediction tasks—and to examine the conditions under which these problems can be effectively addressed—we adapt the thematic analysis framework introduced by~\citet{DBLP:journals/csur/ZhouXTZ21} to suit our research objectives. Specifically, we apply a refined threefold categorization: (i) the prediction type, distinguishing between classification (discrete outcomes) and regression (continuous scores); (ii) the temporal strategy, indicating whether predictions are made before content publication (ex-ante) or after (peeking); and (iii) the predictive granularity, which includes micro-level predictions (individual users or posts), meso-level predictions (community- or group-based behaviors), and macro-level predictions (collective or cascade dynamics across the platform). 

\begin{figure}[ht]
  \centering
  \includegraphics[scale=0.4]{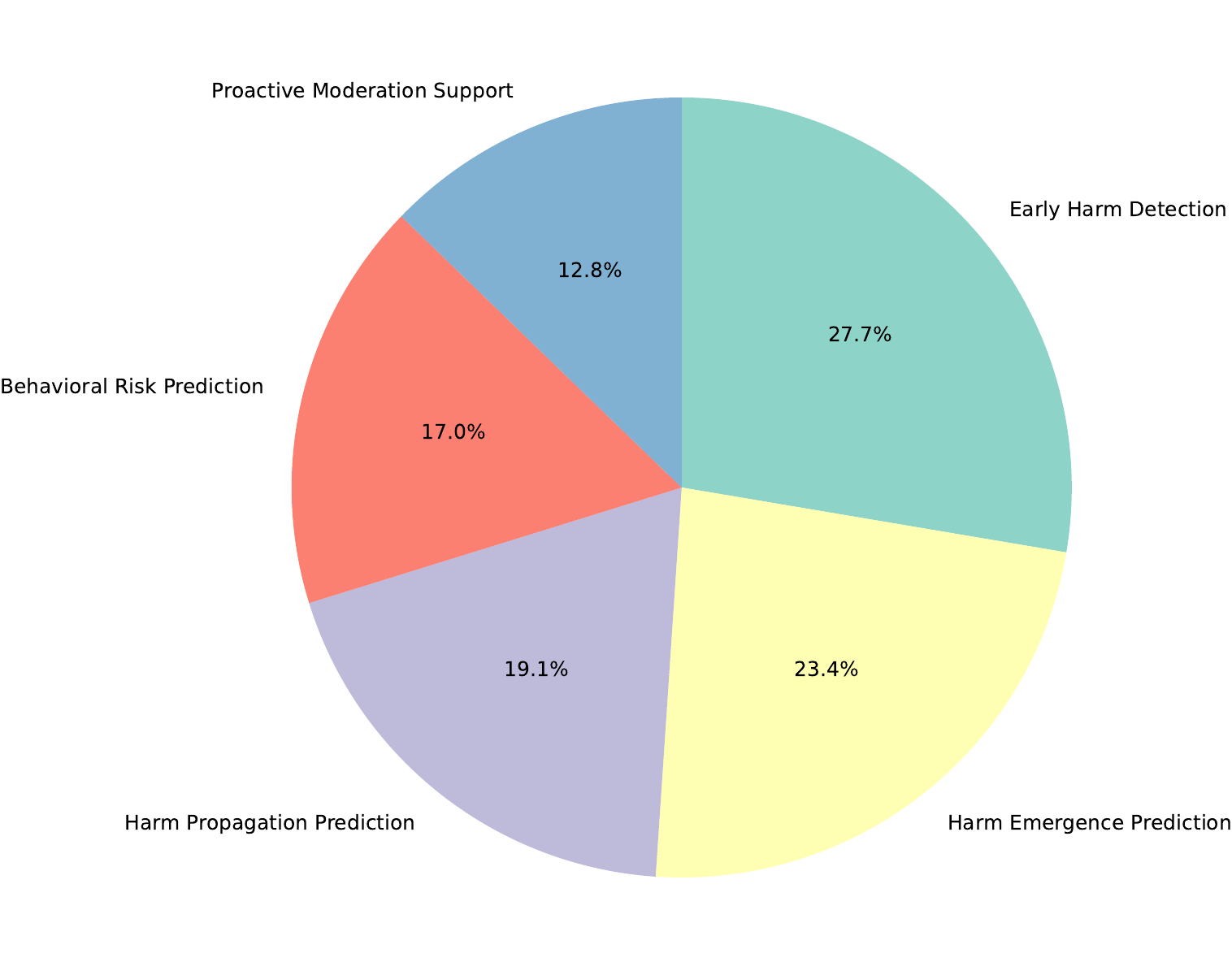}
  \caption{Distribution of ASB prediction task categories across reviewed papers.}
  \label{fig:task_distribution_pie}
\end{figure}

The distribution of task categories in current ASB prediction research, illustrated in Figure~\ref{fig:task_distribution_pie}, reveals a clear concentration around early harm detection and harm emergence prediction, which together represent the majority of studied scenarios. Early detection tasks are designed to anticipate escalation risks in conversations, flag toxic replies at their onset, or detect early signs of conflict based on minimal interaction cues (e.g.,~\cite{almerekhi2020these,mubarak2023detecting,DBLP:conf/sac/KimKY25}). Harm emergence tasks focus on forecasting whether harmful behavior will surface in the course of an interaction, such as predicting conversational derailment or the likelihood of toxic replies (e.g.,~\cite{DBLP:journals/access/NonakaY25,DBLP:journals/corr/abs-2306-12982}). Harm Propagation Prediction forms a substantial secondary cluster, encompassing tasks like modeling the spread and virality of hate speech, estimating future hate intensity, or forecasting how toxic content diffuses across social networks~\cite{makkar2020hate,meng2023predicting,DBLP:journals/corr/abs-2503-03005}. In parallel, behavioral risk prediction—though less frequent—covers user-centric forecasting, such as predicting recidivism, moderation outcomes, or the likelihood of becoming a hate speech amplifier~\cite{DBLP:conf/icdm/PothaM14,an2021predicting,khan2024predicting}. Lastly, proactive moderation support includes predictive tasks aimed at assisting moderation efforts in advance, such as estimating abuse likelihood at posting time, predicting conversation resilience, or evaluating harassment severity through multimodal features~\cite{talukder2018abusniff,bao2021conversations,lambert2022conversational}. Overall, this landscape reflects a multidimensional field, with strong emphasis on early content-level interventions and growing attention toward user trajectories and longitudinal harm forecasting.

\paragraph{Classification versus Regression.}

In ASB prediction research, problem formulation generally falls into two categories: classification and regression, depending on whether the task involves predicting discrete outcomes or continuous values. Classification tasks dominate the field and are typically concerned with determining whether harmful or risky behavior will occur, often as binary outcomes (e.g., toxic vs. non-toxic~\cite{DBLP:journals/corr/abs-2009-10277,DBLP:conf/www/SaveskiRR21}, banned vs. not banned~\cite{DBLP:conf/icwsm/ChengDL15}). Multi-class classification also appears, such as in predicting levels of incivility (e.g., high/medium/low~\cite{DBLP:conf/icwsm/Yu0H24}).

Regression tasks, by contrast, aim to forecast the magnitude or intensity of harm or risk using continuous values. These formulations allow for more nuanced insights, especially in contexts where the severity of ASB evolves over time. Notable examples include forecasting harassment intensity on Instagram through multi-task regression~\cite{chelmis2019minority}, modeling hate intensity over future tweet threads~\cite{dahiya2021would,meng2023predicting}, or estimating continuous hate crime trends~\cite{han2021american}. These tasks not only capture gradation but are also suited for real-time or fine-grained moderation strategies.

Classification formulations in ASB are often more tractable, requiring fewer features and simpler model architectures~\cite{mubarak2023detecting,DBLP:conf/amcis/FaladeYA24}. They are suitable for threshold-based moderation decisions and are commonly used for early detection and risk flagging. Regression tasks, while offering finer granularity and greater explanatory power, typically demand richer temporal or contextual data and are more sensitive to noise, overfitting, and modeling bias~\cite{han2021american,levy2022understanding}. Nonetheless, the distinction is not always rigid: several ASB prediction tasks—such as hate spread estimation or escalation forecasting—can be flexibly reformulated between classification and regression depending on whether a thresholded or continuous output is desired. This duality reflects both the operational goals of moderation and the complexity of harm as a dynamic phenomenon.

\paragraph{Ex-ante Prediction versus Peeking Strategy.}

A key methodological distinction lies in when the model is allowed to make predictions—this defines whether a system follows an ex-ante strategy or a peeking strategy. At its core, this axis reflects the balance between proactive intervention and prediction accuracy. The ex-ante strategy involves making predictions \textit{at} or \textit{before} the moment a post or interaction is published, without any insight into subsequent user behavior. These models often use static features such as textual content, user metadata, or profile attributes. A common example is predicting whether an Instagram post will lead to cyberbullying based solely on the image, caption, and poster’s social graph~\cite{DBLP:journals/corr/HosseinmardiMRH15a}. A more nuanced variation, termed ex-ante context-aware prediction, leverages richer structural context—such as engagement patterns, follower networks, or personality traits—while still operating before any audience interaction. For instance, some models predict retweet behavior based on personality features and social graph proximity~\cite{han2021american,liu2024news,solovev2023moralized}. 

In contrast, peeking strategies permit models to leverage early interaction signals, such as the first few replies or comments. This additional data often yields more accurate predictions but requires a short delay, making it suitable for early warning systems.  For example, models have been developed to forecast hate intensity in Twitter threads or to detect cyberbullying based on initial comment patterns~\cite{lopez2021early, solovev2023moralized}. A more adaptive variant is sequential or progressive prediction, where models continuously update their predictions as new data arrives and may terminate early once a confidence threshold is met~\cite{almerekhi2020these, meng2023predicting}. This enables flexible moderation pipelines that can adjust the timing of interventions to balance responsiveness with reliability.

As summarized in Table~\ref{tab:strategy-forms}, ex-ante strategies are more prevalent in the literature, particularly for classification tasks~\cite{DBLP:journals/corr/abs-1805-04661, masud2021hate}, due to their appeal in proactive moderation. However, peeking and progressive strategies are gaining traction, especially in multi-turn conversation or cascade-based tasks where the temporal evolution of content is central.

\begin{table}[ht]
\centering
\caption{Strategies and Formulations of Feature-based Models}
\label{tab:strategy-forms}
\begin{tabular}{|l|l|p{5cm}|}
\hline
\textbf{Strategy} & \textbf{Formulation} & \textbf{Reference} \\
\hline
\multirow{3}{*}{Ex-ante} 
& Classification &  \cite{DBLP:journals/corr/HosseinmardiMRH15a,DBLP:conf/asunam/HosseinmardiRHL16,DBLP:journals/corr/abs-1805-04661,talukder2018abusniff, mouheb2019real,makkar2020hate,masud2021hate,bao2021conversations, an2021predicting,irani2021early,mubarak2023detecting,lopez2023site, yuan2023conversation,spann2023predicting,DBLP:journals/corr/abs-2306-12982,DBLP:journals/tnse/EttaCMAPQ24,khan2024predicting,DBLP:conf/amcis/FaladeYA24,DBLP:journals/corr/abs-2404-14846,DBLP:conf/sac/KimKY25,DBLP:journals/corr/abs-2503-02191,DBLP:journals/access/NonakaY25,DBLP:journals/corr/abs-2504-08905,DBLP:conf/naacl/SongPYBH25,DBLP:conf/naacl/SongPYBH25,DBLP:conf/icwsm/HickeyFSLB25} \\
& Regression & \cite{kennedy2020constructing,han2021american,wu2022predicting,levy2022understanding,hebert2022predicting,meng2023predicting,gajo2023identification,DBLP:conf/icwsm/Yu0H24,liu2024news,DBLP:journals/corr/abs-2503-03005} \\
\hline
\multirow{3}{*}{Peeking} 
& Classification & \cite{DBLP:conf/icdm/PothaM14,DBLP:conf/icwsm/ChengDL15, DBLP:conf/www/SaveskiRR21,lopez2021early,lin2021early,lambert2022conversational,solovev2023moralized} \\
& Regression & \cite{chelmis2019minority, dahiya2021would} \\
\hline
\end{tabular}
\end{table}

\paragraph{Macro-, Micro-, and Meso-level.}

The granularity of modeling plays a crucial role in shaping both the scope and application of predictive tasks. Much like research in information cascade forecasting~\cite{DBLP:conf/www/ChengADKL14,DBLP:conf/cikm/CaoSCOC17,DBLP:journals/csur/ZhouXTZ21}, ASB tasks can be divided into micro-, meso-, and macro-level predictions, each reflecting a different resolution of analysis. Micro-level prediction focuses on the behavior or risk status of individual users or content items. For example, tasks such as forecasting whether a specific tweet will be deleted, predicting future user bans, or estimating the likelihood that a given Instagram post will provoke cyberbullying fall into this category~\cite{DBLP:journals/corr/HosseinmardiMRH15a,talukder2018abusniff,DBLP:conf/sac/KimKY25}. These tasks draw on item-level attributes—such as text, images, or user history—and aim to determine the harmfulness or impact of single posts or actors. In contrast, meso-level prediction concerns group dynamics or conversation-level phenomena, operating at the scale of threads, sessions, or communities. Here, the goal is not to evaluate isolated messages, but rather to model interactions among users over time. Representative tasks include predicting whether a Reddit discussion will derail~\cite{DBLP:journals/access/NonakaY25}, assessing resilience within online conversations~\cite{bao2021conversations}, or forecasting hate escalation in the early stages of a session~\cite{almerekhi2020these}. This level captures how interpersonal exchanges evolve into collective harm, making it essential for proactive moderation strategies. Finally, macro-level prediction adopts a global perspective, modeling platform-wide or cascade-level behaviors. These tasks typically involve estimating the overall reach, virality, or societal impact of harmful content~\cite{makkar2020hate, gajo2023identification, liu2024news}. For instance, forecasting long-term hate crime trends, predicting the spread of toxic narratives, or estimating the popularity of hateful posts through diffusion modeling are macro-level endeavors. These approaches often draw on structural network features and aggregate behavioral data.

While micro-level tasks dominate current ASB research due to their precision and actionable outputs, recent literature shows growing interest in meso- and macro-level models, especially for understanding systemic risks and coordinated harms~\cite{hebert2022predicting,gajo2023identification}. Moreover, hybrid approaches that incorporate features from multiple levels—such as using local user dynamics to inform cascade-level predictions~\cite{solovev2023moralized,DBLP:conf/www/SaveskiRR21}—offer promising directions for more robust and context-aware ASB forecasting.

\subsection{RQ2 - Machine Learning Approaches for ASB Prediction}

Feature engineering remains a foundational component across all modeling paradigms—whether the task is framed as classification or regression, executed ex-ante or via peeking, or aimed at micro-, meso-, or macro-level predictions. The effectiveness of predictive models heavily relies on how well features capture the dynamics of harmful interactions, user behaviors, and platform contexts.  In our analysis, we identify four core feature categories commonly used across ASB tasks: content-, behavioral-, contextual-, and temporal-based features. Table~\ref{tab:feature-distri} presents a summary of the frequency and distribution of these feature groups in feature-based models over the past decade.

\begin{table}[ht]
\centering
\caption{Number of papers employing each feature group in feature‑based models (2014–2025)}
\label{tab:feature-distri}
\footnotesize
\setlength{\tabcolsep}{3pt} 
\begin{tabularx}{12.9cm}{l *{12}{>{\centering\arraybackslash}X} >{\centering\arraybackslash}X}
\toprule
\textbf{Feature} & \textbf{2014} & \textbf{2015} & \textbf{2016} & \textbf{2017} & \textbf{2018} & \textbf{2019} & \textbf{2020} & \textbf{2021} & \textbf{2022} & \textbf{2023} & \textbf{2024} & \textbf{2025} & \textbf{Total (\%)} \\
\midrule
Content     & \cellcolor{red!15}1 & \cellcolor{red!15}2 & \cellcolor{red!15}1 & \cellcolor{red!15}1 & \cellcolor{red!15}1 & \cellcolor{red!15}2 & \cellcolor{red!25}3 & \cellcolor{red!50}9 & \cellcolor{red!25}3 & \cellcolor{red!50}9 & \cellcolor{red!25}5 & \cellcolor{red!25}4 & \textbf{41 (27.5\%)} \\
Behavioral  & 0 & \cellcolor{red!15}2 & \cellcolor{red!15}1 & \cellcolor{red!15}1 & \cellcolor{red!15}2 & 0 & \cellcolor{red!15}1 & \cellcolor{red!25}2 & \cellcolor{red!15}1 & \cellcolor{red!25}2 & \cellcolor{red!15}1 & \cellcolor{red!25}6 & \textbf{29 (19.5\%)} \\
Contextual  & 0 & \cellcolor{red!15}2 & \cellcolor{red!15}1 & \cellcolor{red!15}1 & \cellcolor{red!15}2 & 0 & \cellcolor{red!15}1 & \cellcolor{red!25}5 & \cellcolor{red!15}2 & \cellcolor{red!25}6 & \cellcolor{red!15}3 & \cellcolor{red!25}7 & \textbf{30 (20.1\%)} \\
Temporal    & \cellcolor{red!15}1 & \cellcolor{red!15}2 & \cellcolor{red!15}1 & \cellcolor{red!15}1 & 0 & \cellcolor{red!15}1 & \cellcolor{red!15}1 & \cellcolor{red!50}8 & \cellcolor{red!25}3 & \cellcolor{red!25}3 & \cellcolor{red!25}2 & \cellcolor{red!25}6 & \textbf{29 (19.5\%)} \\
\midrule
\textbf{\% by year} 
& 5.4\% & 8.7\% & 5.4\% & 5.4\% & 4.0\% 
& 4.7\% & 5.4\% & 22.3\% & 10.1\% & 15.4\% & 9.7\% & 12.4\% & \\
\bottomrule
\end{tabularx}
\end{table}

Based on the evidence summarized in Table~\ref{tab:feature-distri}, current feature-based models for ASB prediction reveal a clear evolution toward more holistic and context-aware representations. Content-based features remain the most prevalent, comprising 27.5\% of all usage. These features focus on what users post—capturing lexical cues, sentiment polarity, hate lexicons, profanity markers, and multimodal signals like images. Such features remain foundational across a range of tasks, including cyberbullying detection, toxicity classification, and hate speech intensity prediction. For example, studies like~\citet{irani2021early} and~\citet{yuan2023conversation} use sentiment and stylistic indicators to assess risk escalation, while~\citet{chelmis2019minority} incorporate visual elements for forecasting harassment intensity. However, the dominance of content alone is increasingly being complemented by other feature dimensions. Behavioral-based features, accounting for 19.5\%, reflect how users act over time—capturing for instance posting frequency, escalation patterns, ban history, and longitudinal consistency. These are crucial for modeling user trajectories in behavioral risk prediction, such as forecasting future bans~\cite{DBLP:journals/corr/abs-2404-14846} or identifying recurrent aggressors~\cite{an2021predicting}. Contextual features, making up 20.1\% of usage, capture where and with whom users interact. They reflect relational and situational dynamics like subreddit norms~\cite{khan2024predicting}, exposure to toxic networks~\cite{gajo2023identification}, or engagement feedback. Such features are often key in meso-level analyses, where understanding the sociotechnical environment is critical for proactive moderation or community-wide interventions. Finally, temporal features—also used in 19.5\% of models—describe when actions occur and how they evolve over time. These include interaction timing, account age, session duration, or pre/post-moderation behavior. Importantly, their adoption has surged since 2021, reflecting growing interest in real-time monitoring and early detection strategies. Studies like~\citet{meng2023predicting,liu2024news} leverage temporal signals to predict escalation or harmful content deletion with increased accuracy. 


In terms of algorithmic strategies, ASB prediction systems span a wide range—from classical models to large-scale generative systems. Classical machine learning models, such as logistic regression, decision trees, and support vector machines, remain in use for their interpretability and robustness in low-resource settings. For instance, Hosseinmardi et al.~\cite{DBLP:journals/corr/HosseinmardiMRH15a} use such models to detect abuse based on linguistic and graph-based features. Meanwhile, neural models like MLPs, LSTMs, and CNNs are widely adopted for their capacity to learn nonlinear patterns and sequential dependencies, particularly in threaded conversations and escalation prediction~\cite{mouheb2019real, yuan2023conversation}. The recent surge of pretrained language models (PLMs)—such as BERT, RoBERTa, and GPT—has brought powerful contextual embedding techniques to the field. These models, often fine-tuned for downstream ASB tasks, dominate recent work on toxicity classification, derailment detection, and conversational forecasting~\cite{bao2021conversations, spann2023predicting}. Beyond text, structured models such as hierarchical transformers and graph neural networks capture higher-order dependencies in conversations or social networks. These are particularly useful in macro- and meso-level modeling, such as tracking the spread of hate speech~\cite{solovev2023moralized} or identifying community vulnerabilities~\cite{lambert2022conversational}. Furthermore, a growing body of work explores multimodal and generative models, which integrate textual, visual, and network data or simulate user behavior over time. Examples include multimodal transformers for Instagram harassment forecasting~\cite{chelmis2019minority} and GANs for hate speech diffusion modeling~\cite{makkar2020hate}. Finally, hybrid and pipeline systems orchestrate multiple components—such as two-stage classifiers or dynamic peeking strategies—to tackle complex, temporally evolving prediction tasks~\cite{almerekhi2020these, an2021predicting}. These modular approaches are increasingly used in real-world moderation pipelines where different model types must interact efficiently.

Altogether, this landscape of features and model types highlights how ASB prediction has matured into a multidimensional field. It blends traditional and neural methods, interpretable and contextual representations, and static and dynamic learning paradigms—each chosen to meet the demands of specific prediction goals, from real-time intervention to long-term user profiling.


\subsection{RQ3 - Datasets and Sources for ASB Prediction}

\begin{table}[ht]
\centering
\caption{Frequently Used Scenarios in ASB Prediction Literature}
\label{tab:dataset}
\begin{tabularx}{\linewidth}{l l X}
\toprule
\textbf{Platform(s)} & \textbf{Category} & \textbf{Reference(s)} \\
\midrule

\multicolumn{3}{l}{\textbf{Social Networking}} \\
\midrule
Instagram & Social Networking & \cite{DBLP:journals/corr/HosseinmardiMRH15a,DBLP:conf/asunam/HosseinmardiRHL16,chelmis2019minority,lopez2023site} \\
Facebook & Social Networking & \cite{talukder2018abusniff,levy2022understanding} \\
Twitter/X & Social Networking & \cite{DBLP:journals/corr/abs-1805-04661, mouheb2019real, kennedy2020constructing, masud2021hate, DBLP:conf/www/SaveskiRR21,an2021predicting, dahiya2021would,lin2021early, irani2021early,wu2022predicting,meng2023predicting,solovev2023moralized,mubarak2023detecting,DBLP:journals/tnse/EttaCMAPQ24,khan2024predicting, DBLP:journals/corr/abs-2503-03005} \\

\midrule
\multicolumn{3}{l}{\textbf{Discussion and Community Forums}} \\
\midrule
Reddit (incl. CMV) & Discussion Forum & \cite{tsantarliotis2017defining, kennedy2020constructing, almerekhi2020these, bao2021conversations,lambert2022conversational, hebert2022predicting, yuan2023conversation,spann2023predicting,DBLP:journals/corr/abs-2306-12982,DBLP:conf/icwsm/Yu0H24,DBLP:conf/amcis/FaladeYA24, DBLP:journals/corr/abs-2404-14846,DBLP:journals/access/NonakaY25,DBLP:journals/corr/abs-2504-08905,DBLP:conf/naacl/SongPYBH25,DBLP:conf/icwsm/HickeyFSLB25} \\
Wikipedia Talk Pages & Discussion Forum & \cite{yuan2023conversation,DBLP:journals/corr/abs-2306-12982,DBLP:journals/access/NonakaY25,DBLP:journals/corr/abs-2504-08905} \\
Incels.is, Il forum dei brutti & Niche Forums & \cite{gajo2023identification} \\

\midrule
\multicolumn{3}{l}{\textbf{Media and Content Sharing Platforms}} \\
\midrule
YouTube, IGN, Vine & Video Sharing & \cite{DBLP:conf/icwsm/ChengDL15, kennedy2020constructing, lopez2021early,lopez2023site,DBLP:journals/tnse/EttaCMAPQ24} \\
CNN, Breitbart, Patch, NYT & News Platforms & \cite{DBLP:conf/icwsm/ChengDL15,han2021american, liu2024news} \\

\midrule
\multicolumn{3}{l}{\textbf{Other Communication Platforms}} \\
\midrule
-- & Chat Platforms & \cite{DBLP:conf/icdm/PothaM14,DBLP:conf/sac/KimKY25} \\
GitHub (Issues/PRs) & Collaboration Platform & \cite{DBLP:journals/corr/abs-2503-02191} \\
\bottomrule
\end{tabularx}
\end{table}

Table~\ref{tab:dataset} offers a comprehensive overview of datasets commonly employed in the development and evaluation of ASB prediction models across diverse online platforms. These datasets originate from a broad array of sources, including social networking sites (e.g., Twitter, Instagram, Facebook), discussion forums (e.g., Reddit, Wikipedia Talk Pages), media-sharing platforms (e.g., YouTube, Vine, and news comment sections), and other communication tools (e.g., chat logs and collaboration platforms like GitHub). Among them, Twitter and Reddit dominate the landscape, appearing in approximately 24 and 13 studies respectively, highlighting their central role as testbeds for a range of ASB prediction tasks.

\begin{figure}[ht]
  \centering
  \includegraphics[scale=0.6]{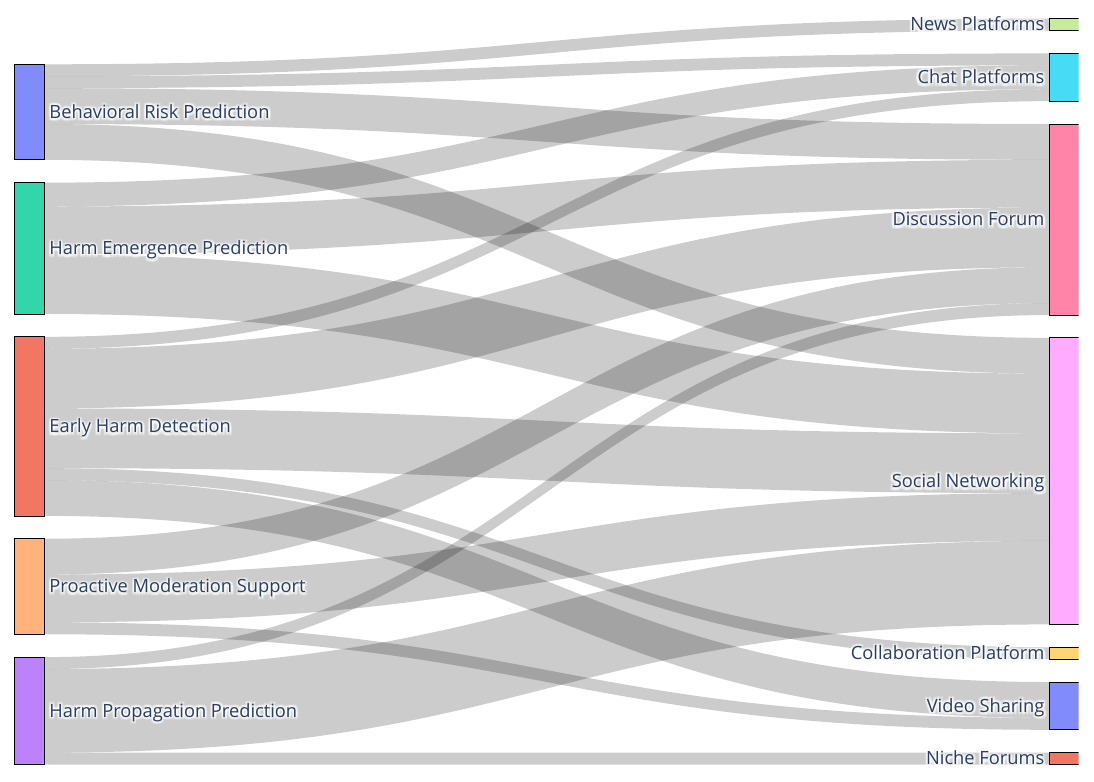}
  \caption{Task–Platform Alignment in ASB Prediction: Chord diagram illustrating the distribution of ASB prediction tasks across different social media platforms.}
  \label{fig:task-platform-chord}
\end{figure}

From an interactional standpoint, datasets used in ASB prediction can be grouped into four structural ``shapes'', each aligned with different modeling needs. (1) Isolated content refers to settings where predictions are made on standalone posts without interaction history—such as individual tweets or toxic news comments~\cite{DBLP:journals/corr/abs-1805-04661,liu2024news}. (2) Local interaction captures short-lived engagement patterns, such as immediate replies or audience responses, often seen in bullying detection tasks within Instagram comment threads~\cite{lopez2021early}. (3) Conversational flow encompasses datasets with multi-turn, threaded interactions—such as Reddit discussions~\cite{lambert2022conversational,DBLP:conf/amcis/FaladeYA24} or Wikipedia Talk Pages~\cite{DBLP:journals/corr/abs-2306-12982,DBLP:journals/access/NonakaY25}—making them ideal for sequential, context-aware modeling of discourse evolution. (4) Networked spread datasets model how harmful content disseminates through user networks, such as retweet cascades on Twitter~\cite{makkar2020hate,masud2021hate}, and are critical for studying diffusion and influence dynamics in ASB contexts. Notably, over half of the surveyed datasets exhibit conversational flow, underscoring a strong focus in the literature on dialogic and escalation-aware prediction tasks. Meanwhile, networked spread structures—used in at least 17 studies—highlight the increasing interest in modeling the propagation and amplification of harmful content across social platforms.

These structural dataset ``shapes'' map closely onto platform affordances, suggesting that the nature of user interaction within each platform plays a key role in determining its suitability for specific ASB prediction tasks. A deeper examination of platform-task alignment, as visualized in Figure~\ref{fig:task-platform-chord}, reveals that dataset selection in ASB prediction research is not arbitrary but often shaped by the inherent communicative architecture of the platform. Twitter, prominently connected to harm emergence and harm propagation tasks in the diagram, is leveraged for its real-time, networked communication dynamics. These properties support modeling of phenomena like hate diffusion, virality, and escalation—for example, in tasks such as hate generation and retweet diffusion prediction~\cite{masud2021hate}, estimating hate intensity in threads~\cite{meng2023predicting}, and forecasting hate speech proliferation~\cite{solovev2023moralized}. In contrast, Reddit and Wikipedia Talk Pages show strong associations with early harm detection and behavioral risk prediction, aligning with their threaded, context-rich conversational formats. These platforms support nuanced modeling of interpersonal breakdowns, moderation outcomes, and conversational derailments, as seen in works predicting civil attacks~\cite{DBLP:journals/corr/abs-2306-12982}, discussion derailment~\cite{DBLP:journals/access/NonakaY25}, and user abandonment post-moderation~\cite{DBLP:journals/corr/abs-2404-14846}. While less central in the diagram, Facebook, Instagram, and YouTube still appear in proactive moderation tasks—particularly where multimodal signals like images and text must be analyzed in tandem. For instance, forecasting harassment intensity often relies on integrating visual and linguistic features, as in~\cite{chelmis2019minority}. The chord diagram thus reinforces how platform-specific structures—whether conversational depth, audience interactivity, or content modality—currently guide the design and focus of ASB prediction efforts.

Lastly, an examination of language diversity across ASB prediction datasets reveals a strong concentration of English-language corpora, which make up more than 83\% of the studies surveyed. This trend corresponds with the dominance of English-speaking users on major platforms such as Twitter and Reddit, as well as the widespread availability of English-language data for academic and computational research. Alongside this, a growing number of studies are beginning to explore datasets in other languages, including Arabic~\cite{mubarak2023detecting}, Italian~\cite{DBLP:journals/tnse/EttaCMAPQ24}, and Korean~\cite{DBLP:conf/sac/KimKY25}, as well as bilingual datasets such as English–Spanish~\cite{irani2021early} and English–Italian~\cite{gajo2023identification}. Some efforts also target more complex linguistic scenarios, incorporating transliterated or mixed-script content~\cite{wu2022predicting}, which are common in multilingual online communities.


\section{Open Challenges}
\label{sec:challenges}

Despite notable progress, ASB prediction research continues to face critical limitations that constrain generalizability, applicability, and scalability. Drawing from the surveyed literature, we highlight five major open challenges, each pointing to directions where further study is both necessary and promising.

\paragraph{1. Linguistic and Cultural Generalization}

A major bottleneck in current ASB prediction systems is the overwhelming reliance on English-language datasets, which account for over 83\% of all surveyed corpora. While this reflects the prominence of English-speaking platforms like Twitter and Reddit, it introduces serious limitations in modeling harmful behavior across languages and cultural contexts. Only a small number of studies incorporate multilingual (e.g., Arabic~\cite{mubarak2023detecting}, Korean~\cite{DBLP:conf/sac/KimKY25}) or cross-script data~\cite{wu2022predicting}, and very few tackle cultural adaptation. As ASB varies in form and social interpretation across regions, there is an urgent need to invest in diverse datasets and culturally grounded modeling strategies that can improve model fairness and inclusivity globally.

\paragraph{2. Platform Dependency and Cross-Domain Transferability}

Current ASB models are often tightly aligned with the affordances of specific platforms. As shown in Figure~\ref{fig:task-platform-chord}, tasks like harm diffusion are commonly studied on Twitter due to its network structure, while Reddit and Wikipedia Talk Pages are preferred for early detection in threaded conversations. While effective, this task-source alignment limits model portability~\cite{DBLP:journals/peerj-cs/YinZ21}. Cross-platform generalization remains largely unaddressed, with few efforts to validate models across settings with different discourse styles, community norms, or moderation policies. Enabling cross-domain learning and transferability is a key future direction—especially for building scalable and resilient moderation systems.

\paragraph{3. Task Formulation and Benchmarking Gaps}

ASB prediction research spans a diverse set of task types—from early harm detection and behavioral forecasting to cascade modeling and harassment intensity estimation. However, the field suffers from limited standardization in how these tasks are formulated, evaluated, and reported. Underexplored areas such as proactive moderation support (e.g., real-time harassment scoring) receive comparatively little attention, while many models remain focused on short-term content classification rather than longer-term behavioral modeling. The absence of standardized evaluation frameworks—including consistent metrics, labeling schemes, and public leaderboards—further impedes replicability and progress tracking across studies. Encouragingly, other branches of online harm research have begun addressing these limitations through shared tasks. The eRisk Lab~\footnote{\hyperref[]{https://erisk.irlab.org/}} hosts annual challenges on early detection of mental health risks and extremism, promoting timeline-based evaluation and early intervention. The PAN Author Profiling task~\footnote{\hyperref[]{https://pan.webis.de/clef21/pan21-web/author-profiling.html}} benchmarks models for identifying hate speech spreaders using user-level behavioral patterns. These initiatives highlight the value of structured, community-driven evaluation. Similar efforts in ASB prediction would improve standardization, enhance comparability, and drive progress toward deployable, generalizable solutions.

\paragraph{4. Temporal Robustness and Concept Drift}

Harmful behavior online evolves rapidly—through new slang, evasion tactics, or shifting platform norms—posing a significant challenge for ASB prediction systems. Unlike more stable classification domains, ASB is deeply tied to social and cultural context, making it highly susceptible to concept drift and temporal variability. However, most existing models are trained and evaluated on static datasets, assuming that linguistic and behavioral patterns remain consistent over time. This leads to rapid performance decay when models are deployed in real-world settings where harmful behavior adapts continuously. For instance,~\cite{app10124180} shows that model accuracy drops sharply when training data is temporally distant from the test set, especially during event-driven language shifts. Crucially, in the ASB context, failing to detect new forms of abuse or emergent threats (e.g., coded hate, targeted harassment campaigns) can have serious social consequences. Robust ASB prediction thus requires longitudinal datasets, temporal adaptation techniques, and continual learning to remain effective against an ever-changing landscape of harm.

\paragraph{5. Interpretability and Human-in-the-Loop Integration}

Modern ASB prediction systems increasingly adopt sophisticated architectures—such as transformers, graph neural networks, and multimodal fusion models—that achieve strong performance but often at the expense of interpretability~\cite{DBLP:conf/aaai/MathewSYBG021}. This opacity poses significant risks when such models are integrated into moderation pipelines, where automated decisions can directly impact user visibility, content removal, or access privileges. The ``right to explanation'', as codified in regulations like the GDPR~\cite{EU2016GDPR}, underscores the need for systems that can justify their outputs in understandable terms, particularly in sensitive domains like hate speech detection. In ASB prediction, where misclassification can lead to unjust sanctions or unchecked harm, model transparency is not just desirable—it is essential. Emerging approaches that jointly learn to classify harmful content and articulate human-aligned justifications offer a promising direction. Ultimately, enhancing explainability and supporting human-in-the-loop moderation are critical for building trustworthy, accountable systems that balance predictive power with ethical oversight.



Together, these challenges reveal that ASB prediction is not solely a technical problem, but a multidimensional one—requiring interdisciplinary collaboration across ML, social science, platform governance, and ethics. Progress on these fronts will be essential to building adaptive, transparent, and equitable systems capable of addressing online harm at scale.

\section{Conclusion}
\label{sec:conclusion}

In this study, we provided a comprehensive survey of the emerging field of ASB prediction, proposing a taxonomy that categorizes existing research into five core task types. Drawing on an extensive analysis of recent literature, we highlighted representative approaches, methodological trends, and platform-specific practices across each category. We also examined the feature engineering strategies and modeling paradigms that shape the current landscape—ranging from classical machine learning to multimodal and graph-based architectures. Furthermore, we identified key open challenges facing the field, including temporal robustness, cross-platform generalization, and the need for explainable, human-centered systems. Our goal is to offer both a snapshot of current progress and a foundation for future research, providing a structured entry point for scholars, developers, and practitioners working to improve the predictive understanding and mitigation of harmful behavior online.

\section{Limitations}

To an external observer, the number of papers reviewed in this study may appear limited. However, this reflects both the emerging nature of ASB prediction as a distinct research subfield and our deliberately focused inclusion criteria. Specifically, we include only machine learning–oriented studies that present a concrete prediction task—such as classification or regression—accompanied by a dataset, modeling framework, or empirical evaluation. This ensures the review captures contributions that advance the development of predictive methodologies for ASB. Our search draws from peer-reviewed conferences, journals, and preprints indexed in Google Scholar, DBLP, Scopus, IEEE Xplore, SpringerLink, and ScienceDirect, consistent with established survey practices in abusive language research (e.g.,~\cite{DBLP:journals/scientometrics/TontodimammaNSF21,DBLP:journals/ijon/Jahan023}).

A further limitation stems from restricted access to closed-access publications. Several potentially relevant works could not be included due to institutional access constraints, particularly those not mirrored in open-access repositories or preprint archives. As a result, the review may underrepresent research published in subscription-only venues, especially outside computer science outlets with strong open-access norms.

Despite comprehensive keyword-driven queries, it is also possible that relevant studies published under alternative terminologies (e.g., ``harmful content forecasting'', ``online risk modeling'') or in adjacent domains (e.g., HCI, computational social science, security studies) were not captured. To mitigate this, the author—an active researcher in the ASB prediction space—supplemented automated searches with a curated archive of domain-relevant publications to reduce omissions. Nonetheless, as the field matures and diversifies, future reviews would benefit from a broader interdisciplinary scope and more inclusive terminologies to better reflect the full range of predictive work addressing antisocial behaviour online.

\section{Ethical Considerations}

Research on ASB prediction necessarily involves engagement with harmful, toxic, or offensive content. This raises ethical risks for both researchers and system users. First and foremost, annotation and model development involve exposure to disturbing material, which may negatively impact the mental health of annotators and practitioners. We echo the guidelines proposed by~\cite{vidgen-etal-2019-challenges,DBLP:conf/emnlp/KirkBVD22} and urge teams to implement protective measures, such as rotating exposure, mental health support, and clear content warnings.

Moreover, privacy considerations are essential, especially for studies using real user data scraped from social platforms. User consent and anonymization protocols must be carefully designed to avoid re-identification risks, particularly in cases involving marginalized or vulnerable users. In proactive moderation or early-warning deployments, care should be taken to avoid disproportionate targeting or false positives, which may reinforce existing biases or enable surveillance-like practices.

Finally, the growing use of generative and predictive models in real-time moderation systems raises questions of responsibility and accountability. While predictive systems offer promise for reducing online harm, they should not be treated as fully autonomous decision-makers. Human-in-the-loop moderation and transparent model governance remain essential to avoid overreach, hallucination, or harm amplification. As with all high-stakes AI systems, ASB prediction requires careful, iterative alignment with both technical robustness and social responsibility.


\bibliography{sn-bibliography}

\end{document}